\documentclass{article}

\usepackage[preprint,nonatbib]{neurips_2021}

\usepackage[utf8]{inputenc} 
\usepackage[T1]{fontenc}    
\usepackage{hyperref}       
\usepackage{url}            
\usepackage{booktabs}       
\usepackage{amsfonts}       
\usepackage{nicefrac}       
\usepackage{microtype}      
\usepackage{xcolor}         
\usepackage[round]{natbib}
\usepackage{array,multirow}
\usepackage{threeparttable}
\usepackage{graphicx}
\usepackage{fixltx2e}

\newcolumntype{P}[1]{>{\centering\arraybackslash}p{#1}}

\title{Leveraging Sentiment Analysis Knowledge to Solve Emotion Detection Tasks}

\author{%
  Maude Nguyen-Thé \\
  Polytechnique Montréal\\
  \texttt{maude.nguyen-the@polymtl.ca} \\
  \And
  Guillaume-Alexandre Bilodeau \\
  Polytechnique Montréal\\
  \texttt{gabilodeau@polymtl.ca} \\
  \And
  Jan Rockemann \\
  Airudi \\
  \texttt{jan.rockemann@airudi.com} \\
}

\begin{document}

\maketitle

\begin{abstract}
  Identifying and understanding underlying sentiment or emotions in text is a key component of multiple natural language processing applications. While simple polarity sentiment analysis is a well-studied subject, fewer advances have been made in identifying more complex, finer-grained emotions using only textual data. In this paper, we present a Transformer-based model with a Fusion of Adapter layers which leverages knowledge from more simple sentiment analysis tasks to improve the emotion detection task on large scale dataset, such as CMU-MOSEI, using the textual modality only. Results show that our proposed method is competitive with other approaches. We obtained state-of-the-art results for emotion recognition on CMU-MOSEI even while using only the textual modality.
\end{abstract}

\section{Introduction}

Sentiment analysis is a subject that has long interested multiple researchers in the domain of natural language understanding. It is a task which aims to identify sentiment polarity for a given signal, which can be of the audio, visual or textual modality. Emotion recognition is a related task which consists of assigning more fine-grained labels, such as anger, joy, sadness, etc.  

This work focuses on analyzing textual inputs. The ability to recognize the sentiment or emotion behind a given sentence or paragraph can lead to multiple applications, such as empathetic dialogue agents and tools to assess the mental state of a patient.  

While sentiment analysis in the form of assigning polarities (positive, negative, and sometimes neutral) to text data is a task that is often studied and for which adequate results have already been obtained for multiple datasets, identifying finer-grained labels such as specific emotions is still a challenge. In addition to the task complexity, in most datasets available for this task, some emotions are much less represented than others, making the training data unbalanced. To address this issue, the model proposed in this work combines knowledge from less complex tasks and is trained using methods to counteract class imbalance. It is based on a Transformer-based model with a Fusion of Adapter layers to leverage knowledge from the more simple sentiment analysis task.

The results obtained are competitive with state-of-the-art multi-modal models on the CMU-MOSEI dataset \citep{bagher-zadeh-etal-2018-multimodal}, while only utilizing the textual modality. Our main contribution can be formulated as:

\begin{itemize}
    \item We designed a method that capitalizes on both pretrained Transformer language models and knowledge from complementary tasks to improve on the emotion recognition task, whilst using Adapter layers that require less training parameters than the conventional fine-tuning approach and taking into account class imbalance.
\end{itemize}

\section{Prior Works and Background}

There are multiple approaches that have been used to solve text-based sentiment analysis and emotion detection tasks, namely rule-based and machine learning approaches. Rule-based approaches consist of creating grammatical and logical rules to assign emotions and use lexicons to assign emotions or polarities to words. Previous works using this approach include the ones of \citet{Udochukwu2015ARA}, \citet{7379415} and  \citet{Seal2019}.  These methods are limited by the size and contents of the lexicon used and by the ambiguity of some keywords. 

Most recent methods are based on the machine learning approach were the network is trained to learn the relationships between words and emotions. Methods such as those proposed by \citet{abdul-mageed-ungar-2017-emonet}, \citet{DBLP:journals/corr/TangQFL15} and \citet{ma-etal-2019-pkuse}  use recurrent neural networks to solve sentiment analysis tasks to break down sentences and understand the relationship between the succession of words and sentiments or emotions. Since the release of pretrained models, recent works \citep{DBLP:journals/corr/abs-1911-02499, Acheampong2021} have been focused on fine-tuning transformer models, which have consistently outperformed previous methods thanks to the multi-head attention applied on words. To improve previous textual emotion recognition methods, we believe that in addition to transfer learning, multi-task learning and class imbalance should be considered. 

\subsection{Transfer Learning}
Transfer learning is a method where the weights of a model trained on a task are used as starting point to train a model for another task. The use of transfer learning with pretrained models has been, for the past few years, the way to obtain state-of-the-art results for multiple natural language understanding (NLU) tasks. Transformer-based pretrained models such as BERT \citep{DBLP:journals/corr/abs-1810-04805}, RoBERTa \citep{DBLP:journals/corr/abs-1907-11692}, XLNet \citep{DBLP:journals/corr/abs-1906-08237}, etc. have been dominating the field over previously used methods. 

\subsection{Multi-Task Learning}

Multi-task learning is used to train one model to solve multiple tasks instead of fine-tuning separate models. Multiple approaches have been used to solve multi-task learning problems. \citet{DBLP:journals/corr/abs-1901-11504} proposed a Multi-Task Deep Neural Network (MT-DNN) with a shared transformer encoder and task-specific heads. \citet{DBLP:journals/corr/abs-1907-04829} and \citet{DBLP:journals/corr/abs-1904-09482} presented a new training procedure based on knowledge distillation to improve the performances of the MT-DNN. These approaches allow the model to learn a shared representation between all tasks. \cite{DBLP:journals/corr/abs-1902-00751} introduced a new model architecture using task-specific adapter layers and keeping the weights of the pretrained encoder frozen. This method, while preventing task interference and catastrophic forgetting, does not allow to transfer knowledge between tasks. To counter this weakness, \citet{DBLP:journals/corr/abs-2005-00247} proposed AdapterFusion, a way to combine knowledge from multiple adapters. 

\subsection{Class Imbalance}
Class imbalance is a challenge in resolving many artificial intelligence tasks. It occurs when one or multiple classes make up significantly less samples of the data than the majority class or classes, often leading to a poor predictive performance for those minority classes. Classic approaches to this problem include re-sampling minority class samples or weighting the loss function according to class frequency. In the field of computer vision, \citet{lin2018focal} proposed a modified version of the cross-entropy loss called the focal loss to handle imbalance.

\section{Proposed Approach}

To improve over previous methods, we have based our method on transfer learning, multi-task learning and we specifically considered class imbalance. To capitalize on transfer learning, our method is based on a strong language model, BERT \citep{DBLP:journals/corr/abs-1810-04805}. We motivate this choice by the fact that identifying emotion requires a good overall understanding of a language, as captured by BERT. Since, sentiment analysis and emotion detection are closely related, we propose a model that learns to combine knowledge from multiple tasks of that nature. This allows leveraging datasets that are annotated only with sentiment for the emotion detection task. Finally, our model is designed to consider class imbalance.

Our method is described in detail in the following.

\subsection{Model}
\begin{figure}[h]
  \centering
  \includegraphics[width=12cm]{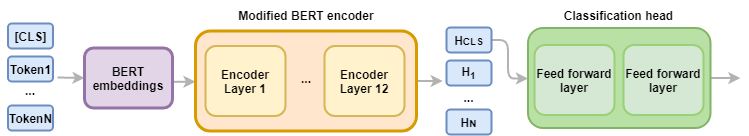}
  \caption{Architecture of the proposed model.}
   \label{fig:wholemodel}
\end{figure}

The proposed model is based on the pretrained Transformer encoder BERT \citep{DBLP:journals/corr/abs-1810-04805} and of a fusion of separately trained Adapter layers. The overall architecture of the model can be seen in Figure \ref{fig:wholemodel}. We chose the BERT encoder (base size), which is comprised of a stack of twelve encoder layers, preceded by token, sentence and position embeddings.  Following the encoder, the last hidden state corresponding to the special classification token ([CLS]) is fed to a classification head formed by two feed forward layers. 

\begin{figure}[h]
  \centering
  \includegraphics[width=10cm]{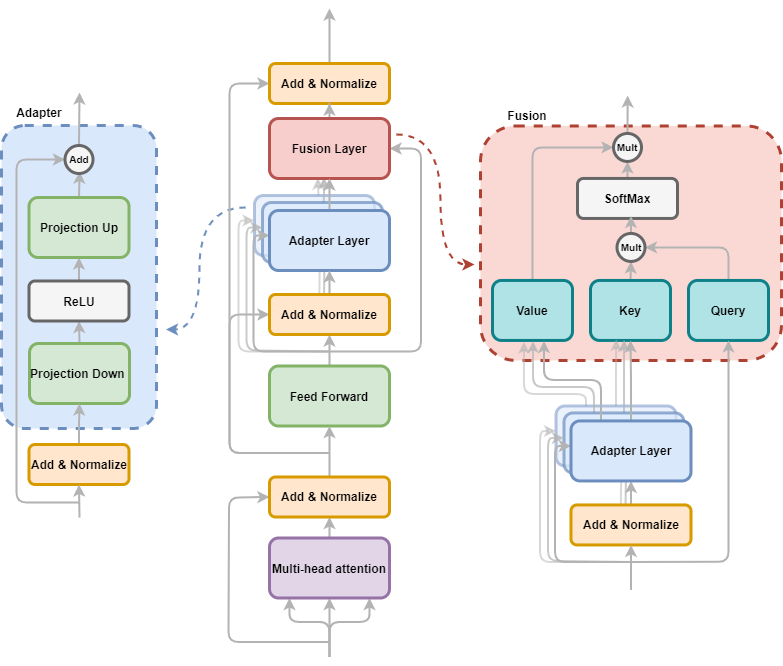}
  \caption{Modified encoder layer proposed in our method (middle), as defined in \citet{DBLP:journals/corr/abs-2005-00247}. On the left, architecture of an adapter layer. On the right, architecture of a fusion layer.}
   \label{fig:adapter}
\end{figure}

Adapters are layers inserted in each of the encoder layers and are trained to adapt the encoder pretrained knowledge to a specific task, while the weights of the encoder are kept frozen (see Figure~\ref{fig:adapter}).  In this work, each adapter layer trained for a specific task has the same structure, which is the one \citet{DBLP:journals/corr/abs-2005-00247} found to be the best across multiple diverse tasks. They are composed of a feed forward layer that projects the encoder hidden state to a lower dimension, a non-linear function and a feed forward layer that projects it back up to the original hidden size. \citet{DBLP:journals/corr/abs-2005-00247} also found that a reduction factor of 16 for the projection down layer adds a reasonable number of parameters per task whilst still achieving good results. All adapters were therefore trained using this reduction factor.

 There are as many adapter layers as there are tasks.  Figure \ref{fig:adapter} illustrates that there are several adapter layers that are used in parallel in our model. To combine the knowledge of each adapter, the AdapterFusion method is used \citep{DBLP:journals/corr/abs-2005-00247}. This method consists of learning a composition of the knowledge of different trained adapters. In this stage of the learning, the weights of the pretrained encoder and of all single adapters are frozen, while the classification and fusion layers are trained. The architecture of the fusion layers is also presented in Figure~\ref{fig:adapter}. 

\subsection{Loss Function}
The loss function used to counter the imbalance present in emotion detection datasets is a modified version of the classic Binary Cross-Entropy (BCE) Loss used for multi-label classification and can be defined as followed:
\begin{equation}
L = \sum_{n=1}^{N} \sum_{c=1}^{C} -w_c [y_{n,c}\log(\sigma(x_{n,c})) + (1-y_{n,c})\log(1-\sigma(x_{n,c}))]
\end{equation}
where $N$ is the number of samples in the batch, $C$ is the number of classes, $x_{n,c}$ is the output of the classification layer of the model for class $c$ of sample $n$, and $w_{c}$ is the positive answer weighting factor for class $c$ defined as:
\[
w_{c} = \frac{\textrm{\# of negative samples of class }c}{\textrm{\# of positive samples of class }c}
\]
This weighting factor is computed on the statistics of the training set data. It weights the loss function to increase recall when the data contains more negative samples of class $c$ than positive samples, and to increase precision in the opposite situation. 

Adapting the focal loss to multi-label classification was also tested but did not significantly improve the performances of the model in comparison to using the classic BCE loss.

\section{Experiments}

Our proposed method was tested using three datasets. We also performed several ablation studies to assess the contribution of each component. 

\subsection{Datasets}
\underline{CMU-MOSEI} \citep{bagher-zadeh-etal-2018-multimodal}: This dataset is comprised of visual, acoustic and textual features for around 23,500 sentences extracted from videos. This dataset is meant to be used to train multi-modal models, but in this work, only the textual inputs were used. The dataset is labelled for sentiment on a scale of [-3,3] and for Ekman emotions \citep{doi:10.1080/02699939208411068} of joy, sadness, anger, surprise, disgust and fear on a scale of [0, 3]. For binary sentiment classification, the labels are binarized to negative (labels lesser than 0) and non-negative (labels greater or equal to 0). The emotions are discretized to non-present (label equal to 0) or present (label greater than 0). Multiple emotions can be present for the same sample. The performance of models on this dataset is measured with standard binary accuracy (A) and F1 scores (F1) for each emotion, as well as an overall non-weighted mean accuracy score and an overall weighted F1 score.

\underline{SST-2} \citep{socher-etal-2013-recursive} \& \underline{IMDB} \citep{maas-etal-2011-learning}: SST-2 is comprised of over 60,000 sentences extracted from movie reviews. IMDB contains 50,000 movie reviews. Both are labelled for sentiment analysis in a 2-class split (positive or negative). These datasets were obtained using the HuggingFace Datasets library\footnote{\url{https://huggingface.co/datasets}}. The performance of models on these datasets is measured with the same binary accuracy scores (A) as CMU-MOSEI.

\subsection{Experimental Setup}
\label{setup}
All experiments use BERT\textsubscript{base} (cased) \citep{DBLP:journals/corr/abs-1810-04805} as the pretrained model, which has 12 encoder layers and a hidden size of 768. Adapter and AdapterFusion layers are added to each of those encoder layers. The classification heads are composed of two fully connected linear layers with sizes equal to the hidden size of the transformer layer (768) and the number of labels (6) respectively, and with $tanh$ activation functions. The input of the first linear layer is the last hidden state of the BERT model corresponding to the classification token ([CLS]) at the beginning of the input sequence.  All models were trained using AdamW \citep{DBLP:journals/corr/abs-1711-05101} with a linear rate scheduler, a learning rate of 1e-5, and a weight decay of 1e-2. All models were trained for 10 epochs with early stopping after 3 epochs if the validation metric did not improve. The Adapter-Transformers library \citep{pfeiffer2020AdapterHub} was used to incorporate the Adapter and AdapterFusion layers to the model. The results presented in the following section are averaged over 3 runs.

Two types of fusion models were trained: one using a fusion of only CMU-MOSEI tasks (Fusion\textsubscript{3}: binary sentiment, 7-class sentiment and emotion classification) and one using additional knowledge from the SST-2 and IMDB sentiment analysis tasks (Fusion\textsubscript{5}). 

\subsection{Results}
The results for the emotion detection task of CMU-MOSEI are presented in Table~\ref{table:resultsMOSEI}. The performance of the proposed model is compared to that of a fine-tuned BERT model and of a model using a single task specific adapter, both using the same classification head as our proposed model. The results of the current state-of-the art model for this dataset \citep{DBLP:journals/corr/abs-2006-15955} are also presented. Note that this state-of-the-art model is a Transformer-based model that utilizes both textual and audio modalities. 

\begin{table}[h]
 \centering
\begin{threeparttable}
  \caption{Results on CMU-MOSEI for emotion detection.  Adapter: BERT with task-specific adapters, Fusion$_3$: BERT with Fusion of adapters for CMU-MOSEI tasks (binary sentiment, 7-class sentiment and emotion classification), Fusion$_5$: BERT with fusion of adapters for tasks of all datasets (CMU-MOSEI tasks and sentiment classification tasks of SST-2 \& IMDB)}
  \label{sample-table}
  \begin{tabular}{c|ccccccc}
    \toprule
    \multirow{3}{*}{\textbf{Model}} & \multicolumn{7}{c}{\textbf{Emotions}} \\

    & Joy & Sadness & Anger & Surprise & Disgust & Fear & Overall \\
    & A/F1 & A/F1 & A/F1 & A/F1 & A/F1 & A/F1 & A/F1 \\
    \midrule
    TBJE\tnote{1} &	66.0/\textbf{71.7}	& \textbf{73.9}/17.8 &	\textbf{81.9}/17.3 &	\textbf{89.2}/3.5	& \textbf{86.5}/45.3 &  \textbf{90.6}/0.0	& \textbf{81.5}/40.5\\
    \midrule
    BERT &	66.3/69.0	& 69.4/42.8 &	74.2/44.3 &	85.8/21.9	& 83.1/\textbf{53.1}	& 83.8/18.7	& 77.1/51.8\\
    Adapter & 67.3/69.4	& 66.3/\textbf{46.1} &	70.4/\textbf{48.5}	& 73.4/26.5	& 77.3/52.3	& 70.9/\textbf{22.7} &	70.9/\textbf{53.7} \\
    Fusion\textsubscript{3} & \textbf{67.5}/70.5 &	66.5/44.4	& 72.5/47.3 & 	81.4/25.9 &	79.0/52.9 &	81.1/21.1 &	74.7/53.6\\
    Fusion\textsubscript{5} & \textbf{67.5}/70.7 &	69.1/44.6 &	73.1/47.5 &	81.3/\textbf{26.6}	& 79.9/53.0 &	82.2/20.3 &	75.5/\textbf{53.7} \\
    \bottomrule
\end{tabular}
\label{table:resultsMOSEI}
\begin{tablenotes}[para, flushleft]
\item[1] Accuracy scores obtained from \citep{DBLP:journals/corr/abs-2006-15955}. F1 scores were computed using the two provided model checkpoints, as the ones presented in their paper were weighted F1 scores.
\end{tablenotes}
\end{threeparttable}
\end{table}

All models trained with our proposed loss function achieve better F1-scores than the current state-of-the-art. While a fully fine-tuned BERT model achieves better overall accuracy, the proposed Fusion model is the one that has best accuracy/F1-score trade-off for all emotions. As observed in Table~\ref{table:stats}, given that all distributions of emotions, except for \textit{joy}, are heavily imbalanced, accuracy is not an appropriate metric for this dataset as it does not fully represent the model ability to identify each emotion. Therefore, it is better to use the F1-score as a measurement basis. Single Adapter models are able to achieve good F1 scores, but do not reach accuracy scores that are comparable to Fusion models, which further proves that combining knowledge from multiple tasks improves the performance of the model. Capitalizing on knowledge from additional sentiment analysis tasks outside of the CMU-MOSEI dataset also allows the Fusion\textsubscript{5} model to perform slightly better than the Fusion\textsubscript{3} model, which only includes knowledge from the CMU-MOSEI tasks. The proposed model also requires a lot less parameters to train, as can be seen in Table~\ref{table:parameters}.

\begin{table}[h]
 \centering
\begin{threeparttable}
  \caption{Positive samples per class for CMU-MOSEI}
  \begin{tabular}{c|cccccc}
    \toprule
    & Joy & Sadness & Anger & Surprise & Disgust & Fear \\
    \midrule
    Proportion of positive samples & ~ 52\%	& ~ 25\% &	~ 21\%	& ~ 10\%	& ~ 17\%	& ~ 8\%  \\
    \bottomrule
\end{tabular}
\label{table:stats}
\begin{tablenotes}[para, flushleft]
\end{tablenotes}
\end{threeparttable}
\end{table}

\begin{table}[h]
 \centering
\begin{threeparttable}
    \caption{Number of parameters per model}
  \begin{tabular}{ccc}
    \toprule
    \textbf{Model} & \textbf{All parameters} & \textbf{Trainable parameters} \\
    \midrule
    BERT (fine-tuned) &	108.3 M	& 108.3 M \\
    Adapter &	109.8 M	& 1.5 M \\
    Fusion$_3$ & 132.8 M &	21.8 M \\
    Fusion$_5$ & 134.6 M & 21.8 M	 \\
    \bottomrule
\end{tabular}
\label{table:parameters}
\end{threeparttable}
\end{table}

\subsection{Comparison of Loss Functions}

The choice of loss function greatly impacts the performance of the model, especially on emotions that are less present in the dataset. The performance obtained with the different loss functions tested are presented in the Table~\ref{table:lossFunctions}.

\begin{table}[h]
 \centering
\begin{threeparttable}
  \caption{Results on CMU-MOSEI for emotion detection for different loss functions. BCE: Regular Binary Cross-Entropy loss, FL: Focal loss, PL: Proposed loss function.}
  \begin{tabular}{c|ccccccc}
    \toprule
    \multirow{3}{*}{\textbf{Loss}} & \multicolumn{7}{c}{\textbf{Emotions}} \\
    & Joy & Sadness & Anger & Surprise & Disgust & Fear & Overall \\
    & A/F1 & A/F1 & A/F1 & A/F1 & A/F1 & A/F1 & A/F1 \\
    \midrule
    BCE & \textbf{67.9}/\textbf{71.5}	& \textbf{75.8}/22.2 &	\textbf{78.5}/25.4	& \textbf{90.5}/1.3	& \textbf{85.6}/48.5	& \textbf{91.7}/0.5 &	\textbf{81.7}/42.8 \\
    FL & 67.7/70.9 &	\textbf{75.8}/24.9	& 78.4/23.1 & 	\textbf{90.5}/0.6 &	85.6/46.1 &	\textbf{91.7}/0.0 &	81.6/42.2\\
    PL & 67.5/70.7 &	69.1/\textbf{44.6} &	73.1/\textbf{47.5} &	81.3/\textbf{26.6}	& 79.9/\textbf{53.0} &	82.2/\textbf{20.3} &	75.5/\textbf{53.7} \\
    \bottomrule
\end{tabular}
\label{table:lossFunctions}

\end{threeparttable}
\end{table}

The difference in performance between the classic Binary Cross-Entropy loss and the Focal loss is not significant. While the use of the loss function proposed in this paper decreases to some extent the accuracy for most emotions, it greatly improves the F1-score for all emotions with the exception of \textit{joy}.


\subsection{Performance of Single Adapters}
\label{adapterPerf}

This section presents the performance of the single Adapters on the other tasks used for knowledge composition in the Fusion models. Tables \ref{table:sent1} and \ref{table:sent2} compare the results obtained by BERT trained with task-specific adapters (Adapter) to fully fine-tuned models and state-of-the-art models. Unless stated otherwise, all accuracy values were obtained by averaging the results over 3 runs using the experimental setup described in section~\ref{setup}. Base size versions of BERT, RoBERTa and XLNet models were used for a fair comparison.

\begin{table}[h]
 \centering
 \caption{Results on CMU-MOSEI sentiment analysis tasks. BERT: fine-tuned BERT, Adapter: BERT with task-specific adapters.}
\begin{threeparttable}
  \begin{tabular}{ccc}
    \toprule
    \multirow{2}{*}{\textbf{Model}} & \textbf{2-class sentiment} & \textbf{7-class sentiment} \\
    & A & A \\
    \midrule
    TBJE\tnote{1} &	84.2	& 45.5 \\
    \midrule
    BERT &	\textbf{84.3} & \textbf{46.8} \\
    Adapter & 83.9 & 46.5  \\
    \bottomrule
\end{tabular}
\label{table:sent1}
\begin{tablenotes}[para, flushleft]
\item[1] Accuracy scores obtained from \citep{DBLP:journals/corr/abs-2006-15955}.
\end{tablenotes}
\end{threeparttable}
\end{table}

The performances of the Adapter models are comparable to those of a fully fine-tuned BERT model. In the case of CMU-MOSEI tasks, they performed on par with or better than state-of-the-art results. In the case of SST-2 and IMDB tasks, they slightly underperformed compared to state-of-the-art fine-tuned language models. However, regardless of the dataset, this experiment shows that adapters can capture useful task-specific information at lower training cost. Furthermore, adapter fusion allows to combine the knowledge from these several good performing task-specific adapters. This explains why our proposed adapter fusion model benefits from the related tasks of sentiment analysis to improve emotion recognition. 

\begin{table}[h]
 \centering
 \caption{Results on SST-2 \& IMDB sentiment analysis tasks. RoBERTa, XLNet and BERT: language model fine-tuned on the specific task, Adapter: BERT with task-specific adapters.}
\begin{threeparttable}
  \begin{tabular}{ccc}
    \toprule
    \multirow{2}{*}{\textbf{Model}} & \textbf{SST-2} & \textbf{IMDB} \\
    & A & A \\
    \midrule
    RoBERTa & \textbf{94.8} & 94.5 \\
    XLNet & 93.4 & \textbf{95.1} \\
    \midrule
    BERT &	93.5 & 94.0 \\
    Adapter & 92.6 & 93.7  \\
    \bottomrule
\end{tabular}
\label{table:sent2}
\end{threeparttable}
\end{table}

\section{Conclusion}

The model presented in this work surpasses state-of-the-art results for emotion recognition on CMU-MOSEI even while using only the textual modality. There is still improvement needed for the rarer emotions in the dataset, but at of the time of producing this article, the results presented are substantially stronger than other contributions in terms of F1-scores. Due to the lack of large-scale datasets for emotion detection in text, testing the model on purely textual data will have to be done in further studies as the data will become available.

\section*{Acknowledgments}
This work was supported by Mitacs through the Mitacs Accelerate program and by Airudi.

\bibliographystyle{plainnat}
\bibliography{references}

\end{document}